\documentclass[10pt,twocolumn,letterpaper]{article}

\usepackage{cvpr}
\usepackage{times}
\usepackage{epsfig}
\usepackage{graphicx}
\usepackage{amsmath}
\usepackage{amssymb}
\usepackage{nopageno}
\usepackage[final]{pdfpages}

\usepackage[pagebackref=true,breaklinks=true,letterpaper=true,colorlinks,bookmarks=false]{hyperref}

\cvprfinalcopy

\def\cvprPaperID{6973} 

\begin{document}

\title{Scalable Uncertainty for Computer Vision with Functional Variational Inference\vspace{-2ex}}

\ifcvprfinal\pagestyle{empty}\fi

\author{Eduardo D C Carvalho\thanks{Authors are with Department of Computing, Imperial College London. Correspondence to {\tt\small eduardo.carvalho16@ic.ac.uk} or  {\tt\small ronald.clark@ic.ac.uk}}\\
\and
Ronald Clark$^*$\\
\and
Andrea Nicastro$^*$\\
\and
Paul H J Kelly$^*$
}

\maketitle

\begin{abstract}
As Deep Learning continues to yield successful applications in Computer Vision, the ability to quantify all forms of uncertainty is a paramount requirement for its safe and reliable deployment in the real-world. In this work, we leverage the formulation of variational inference in function space, where we associate Gaussian Processes (GPs) to both Bayesian CNN priors and variational family. Since GPs are fully determined by their mean and covariance functions, we are able to obtain predictive uncertainty estimates at the cost of a single forward pass through any chosen CNN architecture and for any supervised learning task. By leveraging the structure of the induced covariance matrices, we propose numerically efficient algorithms which enable fast training in the context of high-dimensional tasks such as depth estimation and semantic segmentation. Additionally, we provide sufficient conditions for constructing regression loss functions whose probabilistic counterparts are compatible with aleatoric uncertainty quantification.
\end{abstract}

\vspace{-4mm}

\section{Introduction}

Supervised learning, in its deterministic formulation, involves learning a mapping $f:\mathcal{X} \rightarrow \mathcal{Y}$  given observed data $\mathcal{D}_N=\{x_i, y_i\}_{i=1}^N=\{\boldsymbol{X}_D, \boldsymbol{y}_D\}$. In a Deep Learning context, $f$ is parametrized by a neural network whose architecture expresses convenient inductive biases for the task of interest and whose training consists on optimizing a loss function with respect to its parameters by using stochastic optimization techniques. Despite its widespread empirical success, Deep Learning approaches are hardly ever transparent, so that in certain domains, such as medical diagnosis or self-driving vehicles, it becomes unclear how to map predictions on unseen inputs to a non-catastrophic decision. Thus much research has been focused on obtaining uncertainties from deep models for common computer vision tasks such as semantic segmentation \cite{kampffmeyer2016semantic,huang2018efficient,mukhoti2018evaluating}, depth estimation \cite{kendall2017uncertainties,laidlowtowards}, visual odometry \cite{Bhattacharyya_2018_CVPR,wang2018end,clark2017vinet,clark2017vidloc}, SLAM \cite{czarnowski2020deepfactors} and active learning \cite{gal2017deep}.

The most reliable approach is to consider a Bayesian probabilistic formulation of deep supervised learning, also known as Bayesian Deep Learning \cite{mackay1992practical,neal1996}, so that all forms of predictive uncertainty may be quantified. There are two types of uncertainty one may encounter: $\textit{epistemic}$ and $\textit{aleatoric}$ \cite{kendall2017uncertainties}, both which are naturally accounted for in a Bayesian framework. Epistemic uncertainty is associated with a model's inability of finding a meaningful mapping from inputs to outputs and will eventually vanish as it is trained on a large and diverse dataset. Epistemic uncertainty becomes particularly relevant when the trained model has to make predictions on input examples which, in some sense, differ significantly from training data: out-of-distribution (OOD) inputs \cite{hafner2018ncp}. Aleatoric uncertainty is associated to noise contained in the observed data and cannot be reduced as more data is observed, nor does it increase on OOD inputs, so that it is not able to detect these by itself. Modelling the combination of epistemic and aleatoric uncertainties is therefore key in order to build deep learning based systems which are transparent about their predictive capabilities.

\subsection{General background}

Denoting all parameters of a neural network as $W$, Bayesian Deep Learning starts with positing a prior distribution $\pi(W)$, typically multivariate normal, and a likelihood $p(y|T(x;W))$, where $T(.;W)$ is a neural network with weights $W$. The solution to this bayesian inference problem is the posterior over weights $p(W|\mathcal{D}_N)$, which is unknown due to the intractable computation of marginal likelihood $p(\mathcal{D}_N)$. Stochastic variational inference (SVI) \cite{graves2011practical,hoffman2013stochastic} allows one to perform scalable approximate posterior inference, hence being the dominant paradigm in Bayesian Deep Learning. Denoting $q(W)$ as the variational distribution and $\mathcal{D}_B$ as a mini-batch of size $B$, the following training objective is considered:

\begin{multline}
    \frac{N}{B} \sum_{i=1}^B \mathbb{E}_{q(W)}\left [ \log p(y_i|T(x_i;W))\right] 
    - \mathrm{KL}\left(q(W)||\pi(W)\right)
    \label{elbo_weight_vi}
\end{multline}

This quantity is denoted as evidence lower bound (ELBO), given that it is bounded above by $\log p(\mathcal{D}_N)$. By choosing a convenient family of distributions for $q(W)$ and suitably parametrizing it with neural network mappings, approximate bayesian inference amounts to maximizing the ELBO with respect to its parameters over multiple mini-batches $\mathcal{D}_B$. The success of  variational inference (VI) depends on the expressive capability of $q(W)$, which ideally should be enough to approximate $p(W|\mathcal{D}_N)$. Even though considerable work has been done in designing various variational families for BNN posterior inference \cite{blundell2015weight,louizos2016structured,louizos2017multiplicative,shi2018kernel}, these are not easily applicable in computer vision tasks which require large network architectures.

Alternatively, a nonparametric formulation of probabilistic supervised learning is obtained by introducing a stochastic process over a chosen function space. An $\mathcal{F}$ valued stochastic process with index set $\mathcal{X}$ is a collection of random variables $\{f(x)\}_{x \in \mathcal{X}}$ whose distribution is fully determined by its finite $n$-dimensional marginal distributions $p(f^{\boldsymbol{X}})$, for any $\boldsymbol{X} = (x_1, ..., x_n) \in \mathcal{X}^n$, $n \in \mathbb{N}$, and where $f^{\boldsymbol{X}} = (f(x_1), ..., f(x_n))$. An important class are Gaussian Processes (GPs) \cite{rasmussen2003gaussian}, which are defined by a mean function $m(.)$ and covariance kernel $k(.,.)$, and all its finite dimensional marginal distributions are multivariate gaussians: $p(f^{\boldsymbol{X}}) = \mathcal{N}(m(\boldsymbol{X}), k(\boldsymbol{X}, \boldsymbol{X}))$, where $m(\boldsymbol{X})$ is a mean vector and $k(\boldsymbol{X}, \boldsymbol{X})$ a covariance matrix. 

Bayesian Neural Networks (BNNs) may also be viewed as prior distributions over functions by means of a two-step generative process. Firstly one draws a prior sample $W \sim \pi(W)$, and then a single function is defined by setting $f(.) = T(.;W)$. BNNs are an example of implicit stochastic processes \cite{ma2019variational}, where for any finite set of inputs $\boldsymbol{X}$ its distribution may be written as follows:

\begin{equation}
    p\left(f^{\boldsymbol{X}} \in A\right) = \int_{\{T(\boldsymbol{X};W)=f^{\boldsymbol{X}} \in A\}} \pi(W) dW
\end{equation}

Where $p(.)$ is a probability measure and $A$ is an arbitrary measurable set. Even though it is easy to sample from $p(.)$, it is not generally possible to exactly compute its value due to non-invertibility of $T(.;W)$. Note that in this formulation the dimensionality of the BNN prior does not depend on the dimensionality of weight space, meaning that posterior inference over a BNN with millions of weights only depends on the number of inputs $n$ and dimensionality of $\mathcal{F}$, which is significantly smaller. Moreover, while $p(W|\mathcal{D}_N)$ may have complex structure due to the fact that many different values of $W$ yield the same output values, this can largely be avoided if one performs VI directly in function space \cite{ma2019variational}. 

\subsection{List of contributions}

Our contributions are the following:

\begin{enumerate}
\item Given any loss function of interest for regression tasks, we provide sufficient conditions for constructing well-defined likelihoods which are compatible with aleatoric uncertainty quantification, and provide a practically relevant example based on the reverse Huber loss \cite{lambert2016adaptive,laina2016deeper}.

\item Leveraging the functional VI framework from \cite{sun2018functional}, we propose a computationally scalable variant which uses a suitably parametrized GP as the variational family. Following \cite{aga2018cnngp}, we are able to associate certain Bayesian CNN priors with a closed-form covariance kernel, which we then use to define a GP prior. Assuming the prior is independent across its output dimensions, we propose an efficient method for obtaining its inverse covariance matrix and determinant, hence allowing functional VI to scale to high-dimensional supervised learning tasks. After training, this constitutes a practically useful means of obtaining predictive uncertainty (both epistemic and aleatoric) at the cost of a single forward pass through the network architecture, hence opening new directions for encompassing uncertainty quantification into real-time prediction tasks \cite{kendall2017uncertainties}.

\item We apply this approach in the context of semantic segmentation and depth estimation, where we show it displays well-calibrated uncertainty estimates and error metrics which are comparable with other approaches based on weight-space VI objectives.
\end{enumerate}

\section{Functional Variational Inference}

\subsection{Background}

Even though GPs offer a principled way of handling uncertainty in supervised learning, performing exact inference carries a cubic cost in the number of data points, thus preventing its applicability to large and high-dimensional datasets. Sparse variational methods \cite{titsias2009variational,hensman2013gaussian} overcome this issue by allowing one to compute variational posterior approximations using subsets of training data, but it is difficult to choose an appropriate set of inducing points in the context of image-based datasets \cite{shi2019scalable}. 

Functional Variational Bayesian Neural Networks (FVBNNs) \cite{sun2018functional} use BNNs to approximate function posteriors at finite sets of inputs. This is made possible by defining a KL divergence on general stochastic processes (see \cite{sun2018functional} for the definition and proof). Building upon such divergence, and defining $\boldsymbol{X'} \in \mathcal{X}^{n'}$, where $n'$ is fixed, and setting $\boldsymbol{X} = \boldsymbol{X}_D \cup \boldsymbol{X'}$, it is possible to obtain a practically useful analogue of ELBO in function space:

\begin{equation}
    \sum_{i=1}^N \mathbb{E}_{q(f(x_i))} \left[\log p(y_i|f(x_i)) \right] - \mathrm{KL}\left(q(f^{\boldsymbol{X}})||p(f^{\boldsymbol{X}})\right)
    \label{elbo_fvbnn}
\end{equation}

We refer to this equation as the \textit{functional VI} objective, whose structure will be discussed and simplified during the next sections in order to yield a more computationally feasible version which does not use BNNs as the variational family nor does so explicitly for its prior.

This objective is valid since it is bounded above by $\log p(\mathcal{D}_N)$ for any choice of $\boldsymbol{X'}$ \cite{sun2018functional}. In practice $\mathcal{D}_N$ is replaced by an expectation over a mini-batch $\mathcal{D}_B$, so that the corresponding ELBO is only a lower-bound to $\log p(\mathcal{D}_B)$ and not $\log p(\mathcal{D}_N)$. During training $\boldsymbol{X'}$ may be sampled at random in order to cover the input domain, such as adding gaussian noise to the existing training inputs. Whenever $\boldsymbol{X'}$ are far from training inputs, $q(.)$ will be encouraged to fit the prior process, whereas the data-driven term will dominate on input locations closer to training data. In this way, the question of obtaining reliable predictive uncertainty estimates on OOD inputs gets reduced to choosing a meaningful prior distribution over functions. In this work we will be choosing $p(.)$ to be Bayesian CNNs, which constitute a diverse class of function priors on image space.

\subsection{Logit attenuation for classification in functional VI}

We now consider classification tasks under the functional VI objective (\ref{elbo_fvbnn}), where we assume that $\mathcal{Y} = \{0,1\}^K$, $K$ is the number of distinct classes and $\mathcal{F}=\mathbb{R}^K$. One of the limitations of this objective is that it is not a lower bound to the log-marginal likelihood of the training dataset. When the true function posterior is not in the same class as $q(.)$, there is no guarantee that this procedure will provide reasonable results \cite{shi2019scalable}. We have observed this when we have first tried it in our segmentation experiments, which has caused model training to converge very slowly.

In order to mitigate this issue, we consider the following discrete likelihood under the functional VI framework:

\begin{equation}
    p(y_k|f(x)) = \frac{\mathrm{exp} \left(f_k^{'}(x)\right)}{\sum_{k=1}^K \mathrm{exp} \left(f_k^{'}(x)\right)} 
    \label{boltzmann_likelihood}
\end{equation}

Where $f_k^{'}(x) = f_k(x)/\sigma^2_k(x)$, so that $p(y_k|f(x))$ is a Boltzmann distribution with re-scaled logits, where scale parameter $\sigma^2_k(x)$ weighs its corresponding logit $f_k(x)$. When included into the functional VI objective (\ref{elbo_fvbnn}), this parametrization enables the model to become robust to erroneous class labels contained in the training data, while also avoiding over-regularization from the function prior which may lead to underfitting. This effect of logit attenuation naturally yields a change in aleatoric uncertainty, as measured in entropy. Moreover, we note that each $\sigma^2_k(x)$ is not easily interpretable in terms of inducing higher or smaller aleatoric uncertainty according to its respective magnitude, so that one has to rely on measuring the total predictive uncertainty in terms of the predictive entropy. Additionally, when encompassed into deterministic models or the weight-space ELBO in (\ref{elbo_weight_vi}), re-scaling logits brings no added flexibility.

\section{Functional VI with general regression loss functions}

It is often the case that best-performing non-probabilistic approaches in computer vision tasks not only have carefully crafted network architectures, but also task-specific loss functions which allow one to encode relevant inductive biases.  The most standard examples are the correspondence between gaussian likelihood and $\mathcal{L}_2$ loss, and also between laplacian likelihood and $\mathcal{L}_1$. However, various loss functions of interest are not immediately recognized as being induced by a known probability distribution, so that it would be of practical relevance to start with positing a loss function and then derive its corresponding likelihood model. Given any additive loss function $\ell : \mathcal{Y} \times \mathcal{F} \rightarrow \mathbb{R}_{\geq 0}$, we define its associated likelihood as follows:

\begin{equation}
    p(y|f(x)) = \frac{\mathrm{exp} \left(-\ell(y, f(x)) \right)}{Z}
    \label{gibbs_likelihood}
\end{equation}

This is known as the Gibbs distribution with energy function $\ell$ and temperature parameter set to 1. $Z = \int_{\mathcal{Y}} \mathrm{exp} \left(-\ell(y, f(x)) \right) dy$ is its normalization constant, potentially depending on $f(x)$, which can either be computed analytically or using numerical integration. Any loss function $\ell(.,.)$ for which $Z$ is finite can be made into a likelihood model, hence being consistent with Bayesian reasoning. Moreover, any strictly positive probability density can be represented as in (\ref{gibbs_likelihood}) for some appropriate choice of $\ell$, which follows from the Hammersley-Clifford theorem \cite{besag1974spatial}. In the context of computer vision, typically involving large amounts of labelled and noise-corrupted data, aleatoric uncertainty tends to be the dominant component of predictive uncertainty \cite{kendall2017uncertainties}. This means that, for each task of interest, one needs to restrict from choosing arbitrary likelihoods to the ones which are compatible with modelling this type of uncertainty. In the following subsection we provide a means of doing so for the task of regression.

\subsection{Aleatoric uncertainty for regression}

Without loss of generality, we assume that $\mathcal{Y}=\mathcal{F}=\mathbb{R}$, so that $p(y|f(x))$ is a univariate conditional density. This covers most practical cases of interest, including per-pixel regression tasks such as depth estimation, and simplifies the notation considerably.

In regression tasks, we are typically interested in writing loss functions of the form $\ell(y, f(x)) = \ell\left(\frac{y - f(x)}{\sigma(x)}\right)$, where $f(x)$ and $\sigma(x)$ are location and scale parameters, respectively. Writing $\ell(y)$ as the standardized loss, we define the standard member of its family of Gibbs distributions as $p_0(y) = \frac{1}{Z_0} \mathrm{exp}(-\ell(y))$. Then $p(y|f(x))=\frac{1}{Z} \mathrm{exp}\left(-\ell \left(\frac{y - f(x)}{\sigma(x)}\right)\right)$, where $Z = \sigma(x) Z_0$, defines a valid location-scale family of likelihoods. Moreover, we require its first and second moments to be finite, so that we may compute or approximate means and variances of the predictive distribution. For instance, this excludes using the Cauchy distribution as a likelihood. Substituting into equation \ref{elbo_fvbnn} and ignoring additive constants, we obtain the following training objective:

\begin{multline}
    -\sum_{i=1}^n \left( \mathbb{E}_{q(f(x_i))} \left[\ell\left(\frac{y_i - f(x_i)}{\sigma(x_i)}\right) \right] + \mathrm{log}\left(\sigma(x_i)\right) \right) \\ -  \mathrm{KL}\left(q(f^{\boldsymbol{X}})||p(f^{\boldsymbol{X}})\right)
    \label{elbo_fvbnn_regression_losses}
\end{multline}

Similarly to \cite{kendall2017uncertainties,kendall2017multi}, we interpret each $\sigma(x_i)$ as a loss attenuation factor which may be learned during training and $\mathrm{log}(\sigma(x_i))$ as its regularization component. 

In order to display the practical utility of this loss-based construction, we consider the reverse Huber (berHu) loss from \cite{lambert2016adaptive}, which has previously been considered in \cite{laina2016deeper} for improving monocular depth estimation, and derive its probabilistic counterpart, which we denote as berHu likelihood (see supplementary material).

\section{Scaling Functional VI to high-dimensional tasks}

Various priors of interest in computer vision applications, including Bayesian CNNs, are implicitly defined by probability measures whose value is not directly computable. \cite{sun2018functional} have considered BNNs both as priors and variational family, where the ELBO gradients have been estimated using Stein Spectral Gradient Estimator \cite{shi2018spectral}. However, due to its reliance on estimating intractable quantities from samples, this approach is not viable for computer vision tasks such as depth estimation, semantic segmentation or object classification with large number of classes, all of which display high-dimensional structure in both its inputs and outputs. In order to overcome this issue, we propose to first associate implicit priors with a Reproducing Kernel Hilbert Space (RKHS) and then defining a multi-output GP prior. 

We consider $\mathcal{X} \subseteq \mathbb{R}^d$, where $d = C H W$ pertains to input images having $C$ channels and $H \times W$ resolution, and $\mathcal{F} \subseteq \mathbb{R}^P$, where $P$ is the output dimension depending on the task. For example, $P=H W$ for monocular depth estimation and $P$ equal to the number of distinct classes for object classification. Without loss of generality, we define $p(f(.))$ as a zero-mean multi-output stochastic process on $\mathcal{L}^2(\mathcal{F})$ whose index set is $\mathcal{X}$. Given two images $x_i$ and $x_j$, $K(x_i,x_j):=\int f(x_i)^T f(x_j) dp(f(x_i), f(x_j))$ is the covariance function of the process, which is a $P \times P$ symmetric positive semi-definite matrix for each pair $(x_i,x_j)$. We then posit a GP prior $\hat{p}(f(.))$ with zero mean and covariance function $K(.,.)$, and write its pair-wise joint distribution $\hat{p}(f(x_i), f(x_j))$ as follows:

\begin{equation}
    \begin{pmatrix}
     f(x_i) \\
     f(x_j)
    \end{pmatrix}\sim \mathcal{N}\left(\begin{pmatrix}
    0 \\
    0
    \end{pmatrix},\begin{pmatrix}
     K(x_i, x_i) & K(x_i, x_j) \\
     K(x_i, x_j) & K(x_j, x_j)
   \end{pmatrix}\right).
\end{equation}

Writing the joint multivariate gaussian distribution for a batch of $B>2$ images is straightforward: it is $BP$ dimensional with zero mean vector, and its $BP \times BP$ covariance matrix contains $B^2$ blocks of $P\times P$ matrices, each of which is the evaluation of $K(.,.)$ at the corresponding pair of images. Matrices across the diagonal in the block describe the covariances between pixel locations for each image, whereas the off-diagonal ones describe the correlation between pixel locations of different images.

In the dense case, obtaining the inverse of the full covariance matrix is of complexity $O(B^3 P^3)$ and carries a memory cost of $O(B^2 P^2)$. Even if one is able to choose small $B$ under the functional VI framework, this case would still be intractable for large $P$. A promising way of overcoming this would be to construct prior covariance functions with special structure across the $P$ output dimensions. Recent work done in \cite{aga2018cnngp,novak2019bayesian,yang2019scaling,yang2019wide} has highlighted that Bayesian CNNs do converge to Gaussian Processes as the number of channels of the hidden layers tends to infinity. In cases where activation functions such as relu and tanh are considered, and the architecture does not contain pooling layers, \cite{aga2018cnngp} shows that it is possible to exactly compute a covariance kernel which emulates the same behaviour as the Bayesian CNN, which is denoted as the \textit{equivalent kernel}. In other words, given any Bayesian CNN of this form, in the limit of large number of channels, the function samples they generate come from a zero-mean Gaussian Process given by this covariance function (see \cite{aga2018cnngp} Figure 2 for an example). This covariance kernel can be computed very efficiently at cost which is proportional to a single forward pass through the equivalent CNN architecture with only one channel per layer, which is due to the fact that the resulting GP is independent and identically distributed over the output channels. Moreover, in the absence of pooling layers \cite{novak2019bayesian}, the resulting kernel only contains the variance terms in its diagonal and all pixel-pixel covariances are 0. Thus, given a mini-batch of $B$ input images, the corresponding prior kernel matrix $\boldsymbol{K}$ has only $O(B^2P)$ non-zero entries and can be written in block structure as follows:

\begin{equation}
    \begin{pmatrix}
    K_{1,1} & \cdots & K_{B,1} \\
    \vdots                   & \ddots &             \vdots \\
     K_{B,1} & \cdots & K_{B,B}
    \end{pmatrix}
    \label{kernel_mat}
\end{equation}

Each sub-matrix $K_{i,j}=K(x_i, x_j)$ is diagonal, hence easy to invert and store. Let $\boldsymbol{K}_{:n,:n}$ denote the $nP \times nP$ sub-matrix obtained by indexing from the top-left corner of $\boldsymbol{K}$, where $n=1,...,B$, and consider the following block sub-matrix $\boldsymbol{K}_{:n+1,:n+1}$:

\begin{equation}
    \begin{pmatrix}
    \boldsymbol{K}_{:n,:n} & \boldsymbol{K}_{:n,n+1} \\
     \boldsymbol{K}_{:n,:n+1}^T & K_{n+1,n+1}
    \end{pmatrix}
\end{equation}

Using the block-matrix inversion formula, we may write $\boldsymbol{K}_{:n+1,:n+1}^{-1}$ as follows:

\begin{align}
    \begin{split}
        & \begin{pmatrix}
             \boldsymbol{A}_{:n,:n} & \boldsymbol{B}_{:n,n} \\
             \boldsymbol{B}_{:n,n}^T & S_{n,n}^{-1}
        \end{pmatrix},
        \\
        & \boldsymbol{A}_{:n,:n}=\boldsymbol{K}_{:n,:n}^{-1}(\boldsymbol{I} + \boldsymbol{K}_{:n,n+1} S_{n,n}^{-1} \boldsymbol{K}_{:n,n+1}^T\boldsymbol{K}_{:n,:n}^{-1}),
        \\
        & \boldsymbol{B}_{:n,n}=\boldsymbol{K}_{:n,:n}^{-1} \boldsymbol{K}_{:n,n+1} S_{n,n}^{-1},
        \\
        & S_{n,n} = K_{n+1,n+1} - \boldsymbol{K}_{:n,:n+1}^T \boldsymbol{K}_{:n,:n} \boldsymbol{K}_{:n,n+1}
    \end{split}
\end{align}

Where $S_{n,n}$ is the Schur-complement of $\boldsymbol{K}_{:n+1,n+1}$. This equivalence holds because $\boldsymbol{K}_{:n+1,:n+1}^{-1}$ is invertible if and only if $\boldsymbol{K}_{:n,:n}$ and $S_{n,n}$ are invertible. Starting from $n=1$, $\boldsymbol{K}_{:n+1,:n+1}^{-1}$ can be recursively computed from $\boldsymbol{K}_{:n,:n}^{-1}$, so that we obtain $\boldsymbol{K}^{-1}$ in the last iteration. This algorithm is of complexity $O(B^2 P)$, where $B$ is much smaller than P since it is a batch-size, hence making functional VI applicable in the context of dense prediction tasks such as depth estimation and semantic segmentation. Additionally, the determinant of $\boldsymbol{K}$ may also be obtained efficiently by noting the following recurrence relation \cite{powell2011calculating}:

\begin{equation}
    \mathrm{det}(\boldsymbol{K}_{:n+1,:n+1}) = \mathrm{det}(\boldsymbol{K}_{:n,:n}) \mathrm{det}(S_{n,n})
\end{equation}

By efficiently and stably computing inverse covariance matrices with the same block structure as $\boldsymbol{K}$ and its respective determinants, we are able to replace $p(f^{\boldsymbol{X}})$ in (\ref{elbo_fvbnn}) with the more convenient multi-output GP surrogate $\hat{p}(f^{\boldsymbol{X}})$. In this work we will only consider Bayesian CNN priors without pooling layers, which are most convenient in dense prediction tasks, in order to yield the structural advantages discussed above and leverage the methodology from \cite{aga2018cnngp,novak2019bayesian}. Nevertheless, given any square-integrable stochastic process, it is possible to estimate $K(x_i, x_j)$ using Monte Carlo (MC) sampling and then associating a GP prior with the estimated multi-output covariance function. This has been done in \cite{novak2019bayesian} in order to handle the cases where Bayesian CNN priors do contain pooling layers. Note that any cost involved in computing $\hat{p}(f^{\boldsymbol{X}})$ is only incurred during training.

\begin{figure}[h!]
    \centering
    {{\includegraphics[width=\columnwidth]{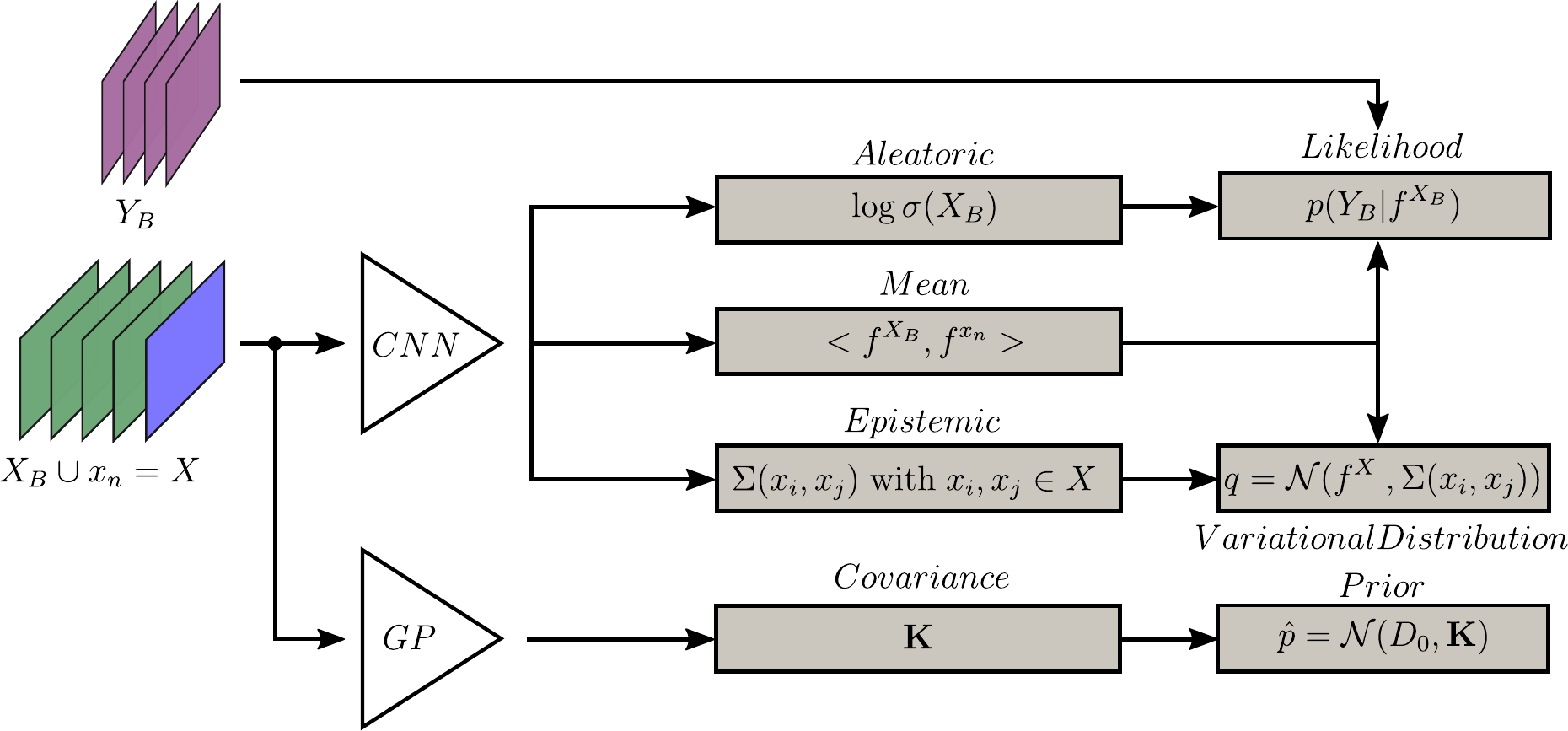} }}%
    \caption{Overview of our functional VI approach. $X_B$ is a batch of rgb inputs, $x_n$ a newly generated one and $D_0$ is the mean function of the GP prior.}%
    \label{fig_fvi}%
\end{figure}

Similarly, by choosing $q(f^{\boldsymbol{X}})$ to be a multi-output GP with mean function $h(.)$ and covariance function $\Sigma(.)$ parametrized by CNN mappings, we are able to compute the corresponding Gaussian KL divergence term in closed form. The expected log-likelihood term may be approximated with MC sampling, but in case of gaussian likelihood it can also be computed in closed form. For each pair of inputs $(x_i, x_j)$, we parametrize the covariance kernel as follows:

\begin{align}
    \Sigma(x_i, x_j)= \frac{1}{L} \sum_{k=1}^L g_k(x_i) \odot g_k(x_j) + D(x_i, x_j) \delta(x_i, x_j)
    \label{q_covariance}
\end{align}

Where each $g_k(x_i), g_k(x_j)$ is a $P$ dimensional feature mapping, $\odot$ denotes the element-wise product and $L<P$, so that the left-term is the diagonal part of a rank-$L$ parameterization. For example, in depth estimation these can be obtained by defining $g(.)$ as a CNN having its output resolution associated with the $P$ pixels and $L$ output channels. $D(x_i,x_j)$ is a diagonal $P \times P$ matrix containing per-pixel variances which is considered only when $x_i=x_j$. This parametrization yields a $P\times P$ diagonal matrix for each pair of inputs, so that the full $BP \times BP$ covariance matrix has the same block structure as in (\ref{kernel_mat}). In this way $q(f^{\boldsymbol{X}})$ is able to account for posterior correlations between different images while being practical to train with mini-batches. Additionally, if one considers regression tasks whose likelihoods are of location-scale family, predictive variances can be computed in closed-form at no additional sampling cost (see supplementary material for an example under the berHu likelihood). In the case of discrete likelihoods, which includes semantic segmentation, computing entropy or mutual-information of the predictive distribution may also be done with a single forward pass plus a small number of gaussian samples, which adds negligible computational cost and is trivially paralellizable. 

In practice, for each input image $x$, we may obtain all quantities of interest as an $R \times (LC + 3C)$ tensor by splitting the output channels of any suitable CNN architecture, where $R$ is the desired output resolution, $C=1$ for tasks such as monocular depth estimation or $C$ equal to the number of classes for tasks such as semantic segmentation. In Figure \ref{fig_fvi} we display a more clear overview of the different components which form our proposed functional VI approach.

\section{Related work}

Monte Carlo Dropout (MCDropout) \cite{gal2016dropout} interprets dropout as positing a variational family in weight-space and uses it at test time in order to compute epistemic uncertainty estimates. MCDropout has since then yielded applications in semantic segmentation tasks \cite{kendall2015bayesian,kampffmeyer2016semantic,kendall2017uncertainties,huang2018efficient,mukhoti2018evaluating}, moncular depth estimation \cite{kendall2017uncertainties}, visual odometry \cite{Bhattacharyya_2018_CVPR} and active learning \cite{gal2017deep}. Despite being convenient to implement during training, the need for multiple forward passes at test time renders MCDropout impractical for both large network architectures (with many dropout layers) and tasks requiring high throughput, such as real-time computer vision. Alternatively, our proposed method allows one to obtain predictive epistemic uncertainty with a single forward pass and to consider a broad range of loss functions whose probabilistic counterparts are consistent with aleatoric uncertainty quantification. 

In the ML literature, various approaches which consider the function space view of BNNs have been discussed in \cite{hafner2018ncp,wang2019function,ma2019variational,pearce2019expressive,khan2019approximate}. Gaussian Process Inference Networks (GPNet) \cite{shi2019scalable} constitutes an alternative to inducing point methods on GPs, and shares some of the motivation of our work in that it also leverages the functional VI objective from \cite{sun2018functional} and chooses both variational family and prior to be GPs. In contrast to any of these, our work focuses on making training and inference practical in the context of dense prediction tasks, which is enabled by suitably parametrizing the variational GP approximation and exploiting special structure in the covariance matrices.

Recently \cite{postels2019sampling} have proposed a scalable method which yields predictive epistemic uncertainty at the cost of a single forward pass. In contrast to it, ours naturally handles all forms of uncertainty, both at training and test times.

\section{Results}

\begin{figure*}[h!]
    \centering
    \qquad{{\includegraphics[width=17cm]{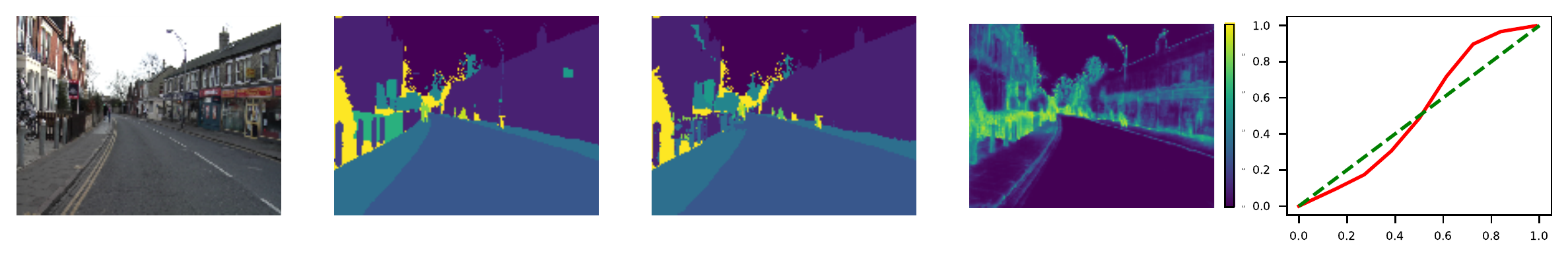} }}%
    \qquad{{\includegraphics[width=17cm]{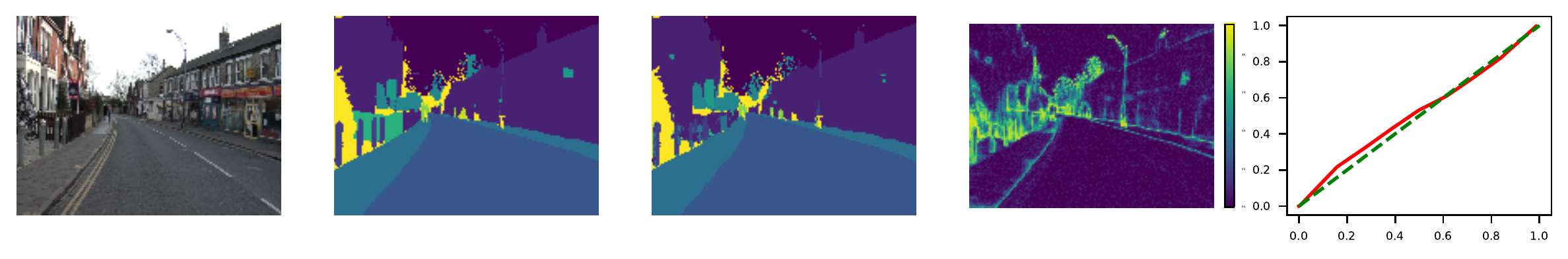} }}%
    \caption{Semantic segmentation on CamVid. MCDropout-Boltzmann (top) and Ours-Boltzmann (bottom). From left to right: rgb input, ground truth, predicted, entropy, calibration plot (as depicted in \cite{kendall2017uncertainties})}%
    \label{fig_camvid}%
    \vspace{-5mm}
\end{figure*}

In order to parametrize the variational GP approximation, we use the FCDenseNet 103 architecture \cite{jegou2017one} without dropout layers. We also adopt this architecture for all other baselines and experiments, using a dropout rate of $0.2$. Even though our initial goal was to closely mimic the setup from \cite{kendall2017uncertainties}, we were not able to reproduce their RMSprop results. Thus, in order to perform a clear comparison, we have decided to compare all methods with the exact same optimizer configurations. For MCDropout, we compute predictions using $S=50$ forward passes at test time.

We choose $L=20$ for the covariance parametrization in (\ref{q_covariance}) and add a constant of $10^{-3}$ to its diagonal during training in order to ensure numerical stability. In order to implement the prior covariance kernel equivalent to a densely connected Bayesian CNN, which has been discussed in section 3, we use the PyTorch implementation made available by the authors in \cite{aga2018cnngp}. For both the segmentation and depth estimation experiments, we compute the equivalent kernel of a densely connected CNN architecture, composed of various convolutions and up-convolutions (see supplementary material), and add a white noise component of variance $0.1$. For the depth experiments, we posit a prior mean of $0.5$ while for segmentation we set it to $1.0$.   In order to generate the inducing inputs $\boldsymbol{X'}$ included in the KL divergence term from equation (\ref{elbo_fvbnn}) during training, we randomly pick one image in the mini-batch and add per-pixel gaussian noise with variance $0.1$.  

\subsection{Semantic Segmentation}

In this section, we consider semantic segmentation on CamVid dataset \cite{brostow2009semantic}. All models have been trained with SGD optimizer, momentum of $0.9$ and weight decay of $10^{-4}$ for $1000$ epochs with batches of size $4$ containing randomly cropped images of resolution $224\times224$, with an initial learning rate of $10^{-3}$ and annealing it every epoch by a factor of $0.998$. Then we finish with training for one epoch on full-sized images with batch size of $1$. We have considered this setup because, while performing our initial experiments by monitoring on the validation set, we have observed that our approach, even though it consistently benefits from fine-tuning on full-sized images in terms of its accuracy measures, the quality of its uncertainty estimates (in terms of calibration score \cite{kuleshov2018accurate}) has degraded significantly. 

For our proposed method, we have used the Boltzmann likelihood with re-scaled logits as given in equation (\ref{boltzmann_likelihood}), which we denote as Ours-Boltzmann. Even though re-scaling logits provides no increase in flexibility to non-functional VI approaches, in order to have the same comparison setup, we chose to parametrize it in the same way for both the deterministic baseline and MCDropout: Deterministic-Boltzmann and MCDropout-Boltzmann, respectively.

From Table \ref{table_camvid} we observe that our method performs best, both in terms of IoU score (averaged over all classes) and accuracy. In Figure \ref{fig_camvid} we display a test example of MCDropout-Boltzmann (top) and Ours-Boltzmann (bottom), where we have masked-out the void class label as yellow. We can see that the uncertainty estimates are reasonable, being higher on segmentation edges and unknown objects. We also include the calibration curve, as computed in \cite{kendall2017uncertainties}, where the green dashed line corresponds to perfect calibration. In order to assess the overall quality of the uncertainty estimates, it is common to compute calibration plots for all pixels in the test set \cite{kendall2017uncertainties,kuleshov2018accurate}. Unfortunately, this is not feasible to compute for our functional VI approach, due to the fact that it captures correlations between multiple images, so that approximating the predictive distribution would require sampling from a high-dimensional non-diagonal gaussian. Thus, in order to enable a simple comparison which works for both Ours-Boltzmann and MCDropout-Boltzmann, we compute the calibration score (see \cite{kuleshov2018accurate}) for each image in the test set and then average, which is given in Table \ref{table_calibration_camvid}. 

\begin{table}[h!]
\centering
\caption{Results from training and testing on CamVid.}
\label{table_camvid}
\scalebox{0.70}{
\begin{tabular}{|l|l|l|l|l|}
\hline
                     & \textbf{IoU} & \textbf{Accuracy} \\ \hline
Deterministic-Boltzmann        &    0.568          &   0.895     \\ \hline
MCDropout-Boltzmann        &     0.556               &    0.893     \\ \hline
Ours-Boltzmann    &   \textbf{0.623}        &     \textbf{0.905}       \\ \hline
\end{tabular}
}
\vspace{-5mm}
\end{table}

\begin{table}[h!]
\centering
\caption{Mean calibration score, computed with 10 equally spaced intervals, averaged over all test set examples. Lower is better.}
\label{table_calibration_camvid}
\scalebox{0.70}{
\begin{tabular}{|l|l|l|l|l|}
\hline
                     & \textbf{Mean Calibration} \\ \hline
MCDropout-Boltzmann        &     0.058       \\ \hline
Ours-Boltzmann    &  \textbf{0.053}     \\ \hline
\end{tabular}
}
\vspace{-5mm}
\end{table}

\subsection{Pixel-wise Depth Regression}

\begin{figure*}[h!]
    \centering
    \qquad{\includegraphics[width=17cm]{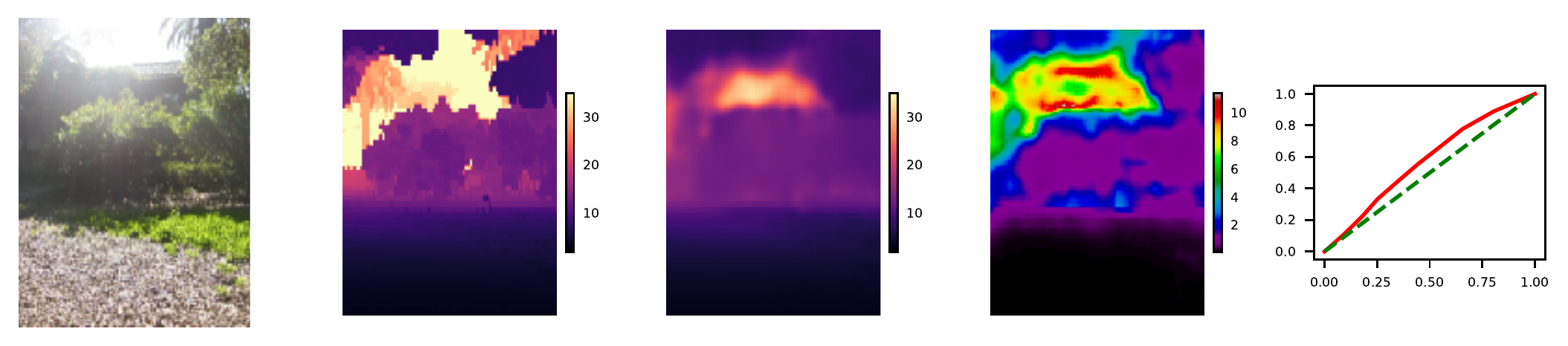}}%
    \qquad{\includegraphics[width=17cm]{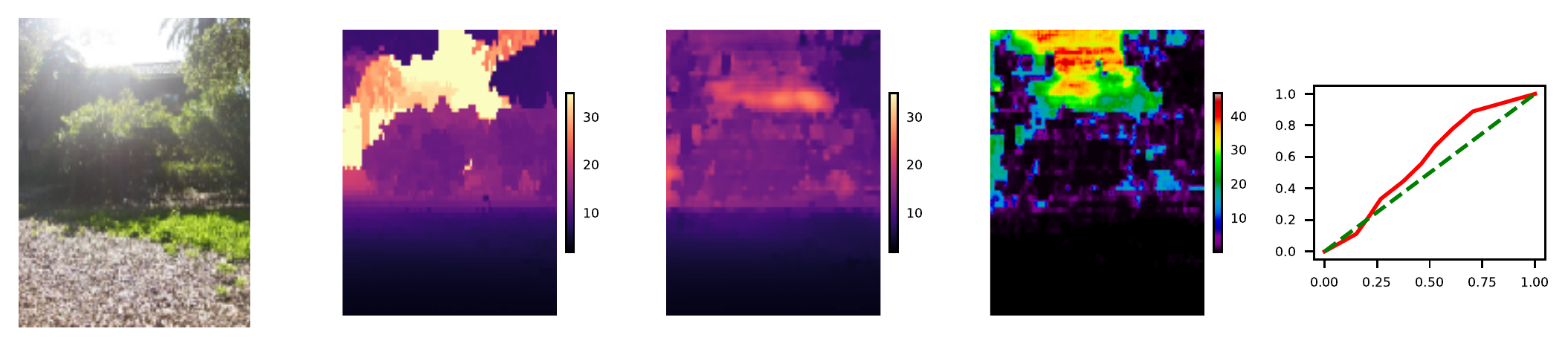}}%
    \caption{Depth estimation on Make3d. MCDropout-Laplace (top) and Ours-Laplace (bottom). From left to right: rgb input, ground truth, predictive mean, predictive standard deviation, calibration plot (as depicted in \cite{kuleshov2018accurate})}%
    \label{fig_depth}%
    \vspace{-5mm}
\end{figure*}

In this section, we consider depth estimation on Make3d dataset \cite{saxena2008make3d}. All models have been trained with AdamW optimizer \cite{adamw2019} with constant learning rate and weight decay set to $10^{-4}$. We have re-sized all images to a resolution of $168\times224$, and trained with a batch size of $4$ for $3000$ epochs. We consider our approach using 3 different likelihoods: Ours-Laplace, Ours-Gaussian and Ours-berHu (as derived in section 3.1.1). We compare with MCDropout-Laplace and two deterministic baselines: Deterministic-$\mathcal{L}_1$ and Deterministic-berHu using the reverse Huber loss \cite{laina2016deeper}.

Test results are displayed in Table \ref{table_make3d}, where MCDropout performs best on all accuracy metrics. To a certain extent, this happened because our proposed method is more sensitive to the choice of batch-size, due to the fact that the functional VI objective is not a lower bound to the log marginal likelihood of the dataset, so that it has underfitted slightly more than MCDropout-Laplace and deterministic methods. Additionally, we had to use a learning rate of $10^{-4}$, as higher values would result in more unstable training for all our functional VI approaches.

In Figure \ref{fig_depth} we plot one test prediction for MCDropout-Laplace (top) and Ours-Laplace (bottom). In this case, we observe one of the benefits of our approach: around the sky area in the image, MCDropout-Laplace is overconfident about its predicted depth map, while ours correctly outputs high predictive uncertainty. Note that this is not reflected in the calibration curves, as all pixels with depth greater than 70m are masked out due to long-range inaccuracies in the dataset \cite{laina2016deeper}.

In Table \ref{table_calibration_make3d} we display the calibration scores for the probabilistic methods (see \cite{kuleshov2018accurate}), averaged over all test images, where Ours-Laplace performs slightly better than MCDropout-Laplace, despite not faring so well in terms of accuracy metrics.

\begin{table}[h!]
\centering
\caption{Results from training and testing on Make3d dataset.}
\label{table_make3d}
\scalebox{0.70}{
\begin{tabular}{|l|l|l|l|}
\hline
                     & \textbf{rel} & \textbf{log10} & \textbf{rms}  \\ \hline
Deterministic-$\mathcal{L}_1$         & 0.212    &     0.085     &           5.29            \\ \hline
Deterministic-berHu        & 0.222   &   0.084       &            5.08           \\ \hline
MCDropout-Laplace      &  \textbf{0.210}  & \textbf{0.081}          &     \textbf{5.05}                  \\ \hline
Ours-Laplace           &   0.264  &      0.092   &   5.74   \\ \hline
Ours-berHu              & 0.237   &    0.088      &   5.68           \\ \hline
Ours-Gaussian             &  0.254  &    0.089      &  5.65            \\ \hline
\end{tabular}
}
\vspace{-5mm}
\end{table}

\begin{table}[h!]
\centering
\caption{Mean calibration score, computed with 10 equally spaced intervals, averaged over all test set examples. Lower is better.}
\label{table_calibration_make3d}
\scalebox{0.70}{
\begin{tabular}{|l|l|l|l|l|}
\hline
                     & \textbf{Mean Calibration} \\ \hline
MCDropout-Laplace        &    0.427        \\ \hline
Ours-Laplace    &   \textbf{0.409}    \\ \hline
Ours-berHu    &   0.631    \\ \hline
Ours-Gaussian    &   0.491    \\ \hline
\end{tabular}
}
\vspace{-5mm}
\end{table}

\subsection{Inference time comparison}

Let $F$ be the inference time of one forward pass from a neural network on a RGB input. Our method's inference time (for obtaining predictive mean and uncertainty) is then $F + c_1$, while for MCDropout is $SF + c_2$, where $c_1,c_2$ are extra time costs needed to obtain the predictive uncertainties. In computer vision $F$ is often the dominant term, since it often involves large network architectures, of which the FCDenseNet 103 architecture is an example. We have tested these claims by performing multiple runs on an NVIDIA RTX6000 GPU, the same device in which all models have been trained and tested. The inference times for depth estimation and semantic segmentation are displayed in Table \ref{table_runtime_make3d} and Table \ref{table_runtime_camvid}, respectively. On depth estimation our method and deterministic had equivalent inference times. On segmentation $c_1$ depends on the number of gaussian samples taken, but is significantly cheaper than $F$ and trivially amenable to parallelization, so that our method still displayed cost of same order as deterministic model. In both cases, MCDropout was approximately $S=50$ times slower than its deterministic counterpart.

\begin{table}[h!]
\centering
\caption{Depth estimation on Make3D. Inference time comparison over 100 independent runs.}
\label{table_runtime_make3d}
\scalebox{0.70}{
\begin{tabular}{|l|l|l|l|l|}
\hline
                     & \textbf{mean $\pm$ std (ms)} \\ \hline
Deterministic-$\mathcal{L}_1$    & 51.29 $\pm$ 1.88 \\ \hline
Deterministic-berHu    &  51.28 $\pm$ 1.62 \\ \hline
MCDropout-Laplace        &      2615.65 $\pm$ 13.75     \\ \hline
Ours-Laplace    &    50.98 $\pm$ 1.74 \\ \hline
Ours-berHu    &    51.43 $\pm$ 2.12  \\ \hline
Ours-Gaussian    &   51.13 $\pm$ 2.20    \\ \hline
\end{tabular}
}
\vspace{-5mm}
\end{table}

\begin{table}[h!]
\centering
\caption{Semantic segmentation on CamVid. Inference time comparison over 100 independent runs.}
\label{table_runtime_camvid}
\scalebox{0.70}{
\begin{tabular}{|l|l|l|l|l|}
\hline
                     & \textbf{mean $\pm$ std (ms)} \\ \hline
Deterministic-Boltzmann    & 111.64 $\pm$ 0.27 \\ \hline
MCDropout-Boltzmann    &   5763.63 $\pm$ 1.95  \\ \hline
Ours-Boltzmann       &   128.59 $\pm$ 1.86   \\
\hline
\end{tabular}
}
\vspace{-5mm}
\end{table}

\section{Conclusion}

We have proposed a method which, by leveraging the functional VI objective from \cite{sun2018functional}, enables efficient training of Bayesian Deep Learning models and whose predictive inference requires only one forward pass, for any supervised learning task and network architecture. This is made possible by replacing the intractable BNN prior by a GP with covariance kernel as derived in \cite{aga2018cnngp}, parametrizing the variational family as a GP with a suitably structured covariance kernel and by leveraging efficient algorithms for matrix inversion and determinant computation during training. Furthermore, we have discussed how to start with a well-defined loss function in regression and then derive its probabilistic counterpart in a way which is consistent with aleatoric uncertainty quantification, having provided the derivation of the berHu likelihood as an example. 

Our framework may readily be applied to other pixel-wise supervised learning tasks. Extending to tasks which benefit from having pooling layers, such as object classification, is also possible but requires some caution. This is because Bayesian CNN priors which contain pooling layers no longer induce GPs which have the special covariance structure displayed in (\ref{kernel_mat}), given that pooling induces local correlations between different pixel locations \cite{novak2019bayesian}.

As a direction of future work, it would be relevant to extend our proposed methodology to account for temporal information. This would be particularly important in monocular depth estimation, which is naturally prone to display high aleatoric uncertainty and would benefit from refined uncertainty estimates over consecutive time-frames \cite{liu2019neural}. Another direction of future work would be to overcome any potential underfitting occurring in pixel-wise regression tasks, as observed in our Make3D depth regression experiment, in which choosing more meaningful function priors and better variational distribution's covariance parametrizations could help. 

\textbf{Acknowledgements}

Eduardo is supported by an EPSRC Industrial CASE scheme in collaboration with Arup. Paul is supported by EPSRC grant reference EP/P010040/1. We would like to thank Jan Czarnowski, Sajad Saeedi, Tristan Laidlow and all our reviewers for helpful insights and comments.

\bibliographystyle{ieee_fullname}

\newpage

\onecolumn



\section*{Supplementary material}

\subsection*{a) Reverse Huber (berHu) likelihood}

The reverse Huber (berHu) loss is defined as follows:

\begin{equation}
    \ell (y) = |y| \mathcal{I}(|y| \leq c) + \left(\frac{y^2 + c^2}{2c}\right) \mathcal{I}(|y| > c)
    \label{berhu_loss}
\end{equation}

Where $\mathcal{I}(.)$ is the indicator function and $c>0$ is an appropriately chosen threshold. $\ell (y)$ yields a balance between $\mathcal{L}_1$ and $\mathcal{L}_2$ losses: for smaller residuals, $\mathcal{L}_1$ is considered in order to yield gradients with larger magnitudes, whereas the $\mathcal{L}_2$ component provides an increased penalty to large residuals so that the network also accounts for these.

In this case, it can be shown that $Z_0 = 2 \left(1 - e^{-\mathrm{c}} + e^{-\mathrm{c}/2} (2 \pi \mathrm{c})^{1/2} \Phi(-c^{1/2})\right)$, where $\Phi(.)$ is the standard normal CDF, and taking $c \rightarrow \infty$ recovers the Laplace distribution. Given a new input $x^*$, the predictive mean $m(x^*)$ and variance $\mathbb{V}(x^*)$ can be written as follows:

\begin{align}
    \label{berhu_pred}
    \begin{split}
        & m(x^*) = m \left( q(f(x^*)) \right) ,
        \\
        & \mathbb{V}(x^*) = \underbrace{w(c) \sigma^2(x^*)}_{\text{aleatoric}} + \underbrace{\mathbb{V}(q(f(x^*))}_{\text{epistemic}} ,
        \\
        & w(\mathrm{c}) = \frac{-4 (\mathrm{c} + 1) e^{-\mathrm{c}} + 4 + 2 e^{-\mathrm{c}/2} (2 \pi)^{1/2} (\mathrm{c})^{3/2} \Phi(-c^{1/2})}{Z_0}
     \end{split}
\end{align}

$m(x^*)$ is the mean of the variational distribution $q(f(x^*))$, which follows from the law of conditional expectation, and the decomposition of its variance as sum of epismetic and aleatoric components follows from the law of total variance. Compared to choosing gaussian or laplacian likelihoods, berHu yields a weighted version of $\sigma^2(x^*)$ which depends on the choice of $c$. We choose $c = \frac{1}{5} \mathrm{max}_i \mathbb{E}_{q(f(x_i))} \left( |y_i - f(x_i)| \right)$ during training, where $i$ indexes all output feature maps in a mini-batch and the inner expectation is replaced by a monte carlo estimate. In order to compute the predictive distribution at test time, we record the maximum value of $c$ across all batches in the final training epoch. 

\section*{b) Bayesian CNN GP prior}

\begin{figure}[h!]
    \centering
    \includegraphics[width=0.5\textwidth]{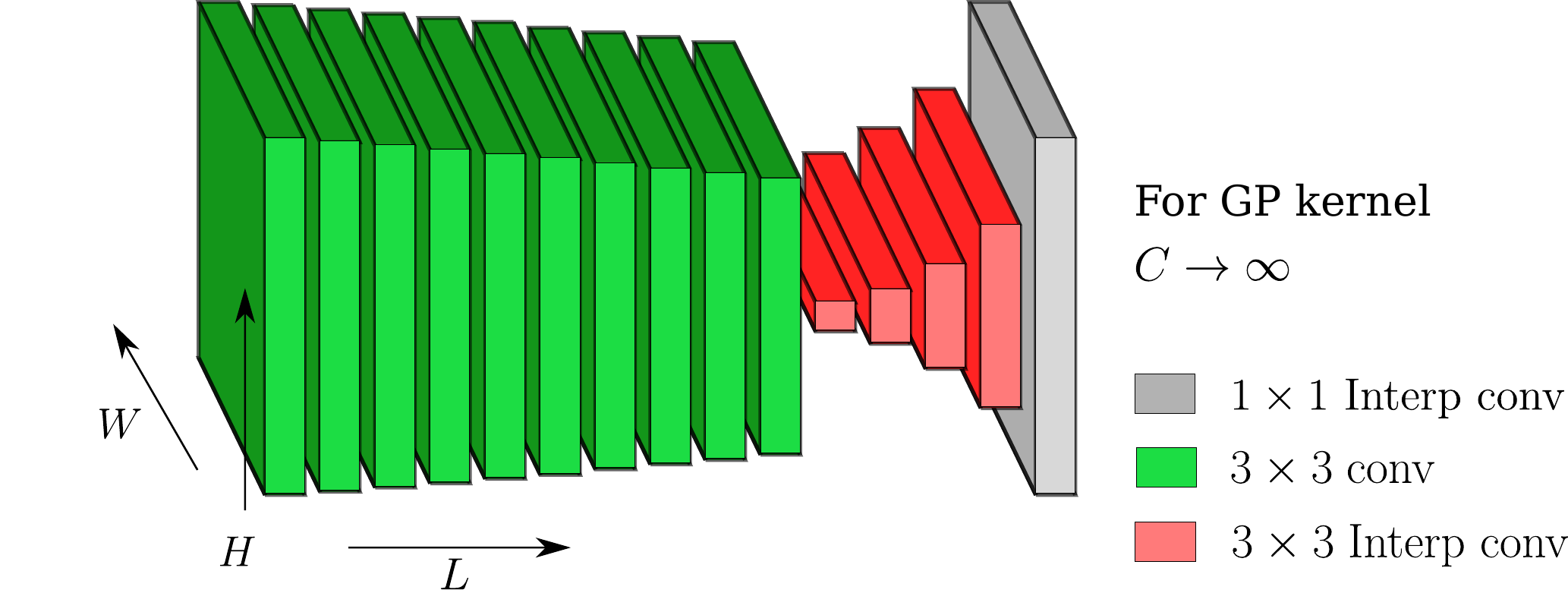}
    \caption{Architecture of the Bayesian CNN GP prior.}
    \label{fig:prior}
\end{figure}

In Figure \ref{fig:prior} we display the Bayesian CNN architecture, with weight prior variance of $0.2$ and bias' prior variance of $0.08$, using relu activations, from which the equivalent kernel was derived. This covariance kernel encompasses the behaviour of this Bayesian CNN architecture in the limit where the number of channels in its hidden layers, $C$, tends to infinity, which we denote as Bayesian CNN GP prior. The red and gray blocks correspond to linear interpolation followed by a convolution layer. The sequence of output resolutions for linear interpolation are $20,40,60,80,100$ percent of the desired output resolution. 

\section*{c) Semantic segmentation on CamVid}

During training, for all methods, we augment the CamVid dataset by performing random horizontal flips with $0.5$. For our approach (Ours-Boltzmann) we estimate the expected log-likelihood term using $20$ monte carlo samples from the variational distribution.

\begin{figure*}[h!]
    \centering
    \hspace{-7mm}\qquad{{\includegraphics[width=15cm]{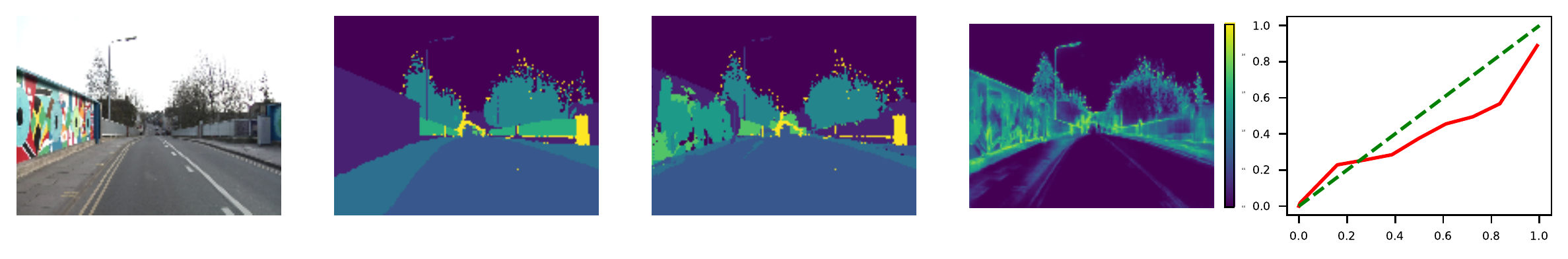}}}
    \qquad{{\includegraphics[width=15cm]{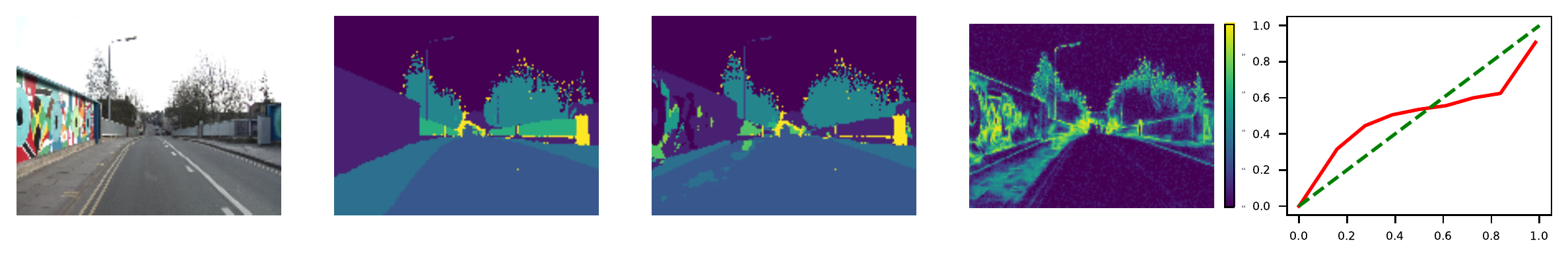} }}%
    \caption{Semantic segmentation on CamVid. MCDropout-Boltzmann (top) and Ours-Boltzmann (bottom). From left to right: rgb input, ground truth, predicted, entropy, calibration plot}%
    \vspace{-5mm}
    \label{fig_seg}
\end{figure*}

\begin{figure*}[h!]
    \centering
    \hspace{-7mm}\qquad{{\includegraphics[width=15cm]{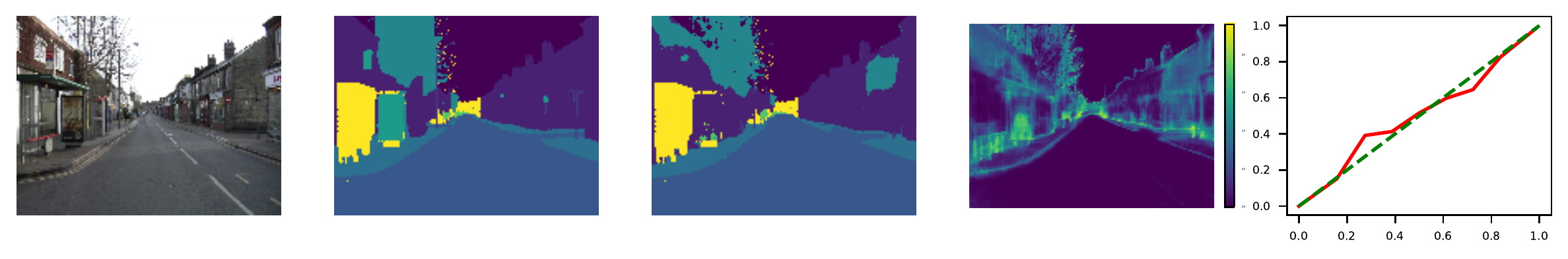}}}
    \qquad{{\includegraphics[width=15cm]{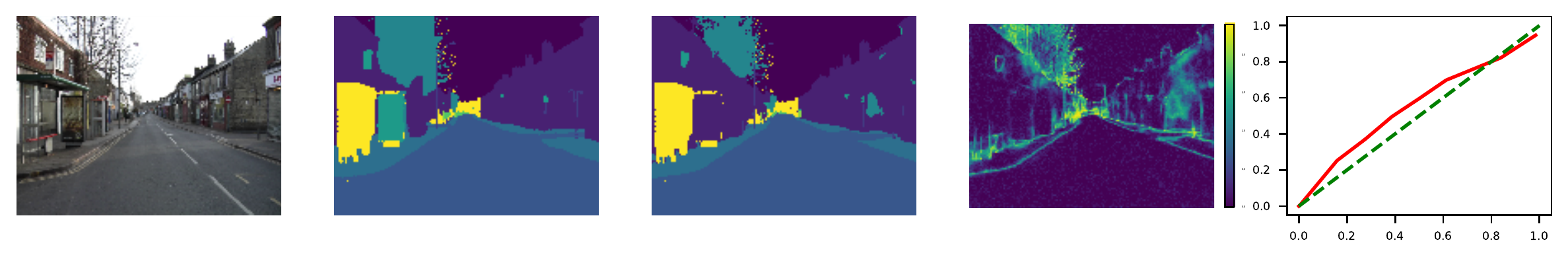} }}%
    \caption{Semantic segmentation on CamVid. MCDropout-Boltzmann (top) and Ours-Boltzmann (bottom). From left to right: rgb input, ground truth, predicted, entropy, calibration plot}%
    \vspace{-5mm}
    \label{fig_seg_failure}
\end{figure*}

We select and discuss two test cases in order to compare our method (Ours-Boltzmann) and MCDropout-Boltzmann. Unknown segmentation classes have been masked out as yellow in the plots corresponding to ground truth and predicted classes. In Figure \ref{fig_seg}, MCDropout-Boltzmann (top) is wrongly overconfident that the left sidewalk is part of the road, while ours correctly accounts for this difference by outputting higher predictive entropy. Figure \ref{fig_seg_failure} displays a failure test case from our method, in which it displays high confidence (low-entropy) that the bus-stop is part of the housing lots. MCDropout-Boltzmann does a better job at flagging out this unknown segmentation class by outputting higher entropy on several of its regions.

\section*{d) Depth estimation on Make3d}

During training, for all methods, we augment the Make3d dataset with random horizontal flips (with probability $0.5$), and randomly adjust brightness, saturation, contrast and hue of rgb inputs by a factor of $0.1$. For Ours-Laplace and Ours-berHu, we estimate the expected log-likelihood term using $50$ monte carlo samples from the variational distribution.

We select and discuss two test cases in order to compare our methods (Ours-Laplace, Ours-Gaussian and Ours-berHu) with MCDropout-Laplace. In Figure \ref{fig_depth} we display an example where all methods perform well in terms of the predicted depth map. We can observe that both the predicted depth maps and uncertainty from our methods have a sharper aspect than MCDropout-Laplace, which we have consistently observed for most predictions. In Figure \ref{fig_depth_failure} we display a failure case for all methods, in terms of predicting inaccurate depth maps. Predictive uncertainty, both its epistemic and aleatoric components, is expected to be higher around the blue sky region. Our methods deliver this effect, while MCDropout-Laplace is overconfident about the predicted depth maps in this region. 

\begin{figure*}[h!]
    \centering
    \hspace{-7mm}\qquad{{\includegraphics[width=15cm]{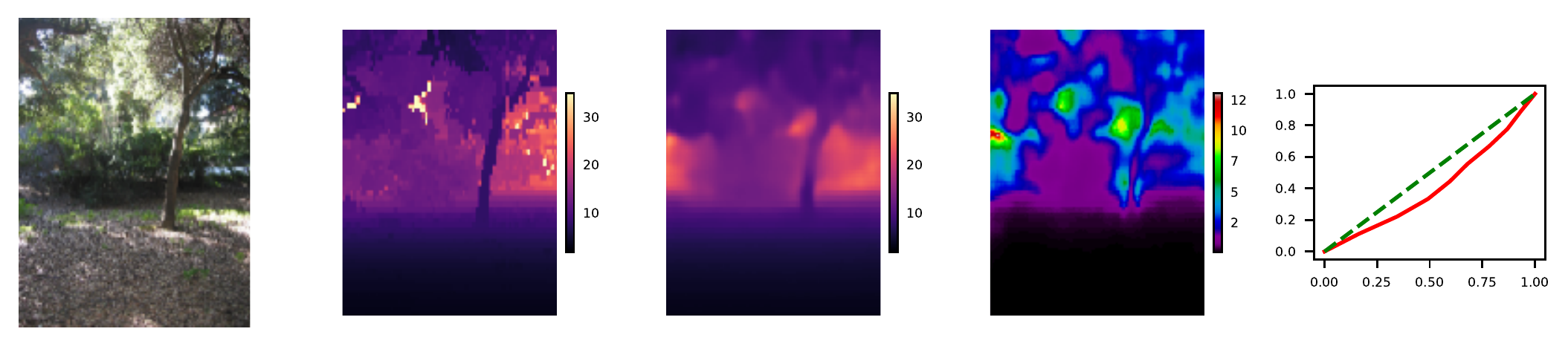} }}%
    \qquad{{\includegraphics[width=15cm]{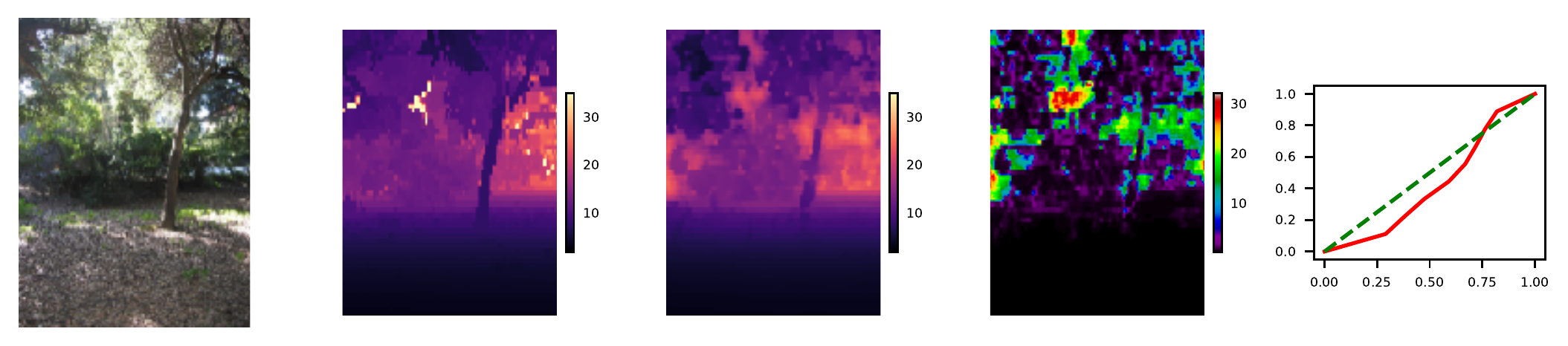} }}%
    \qquad{{\includegraphics[width=15cm]{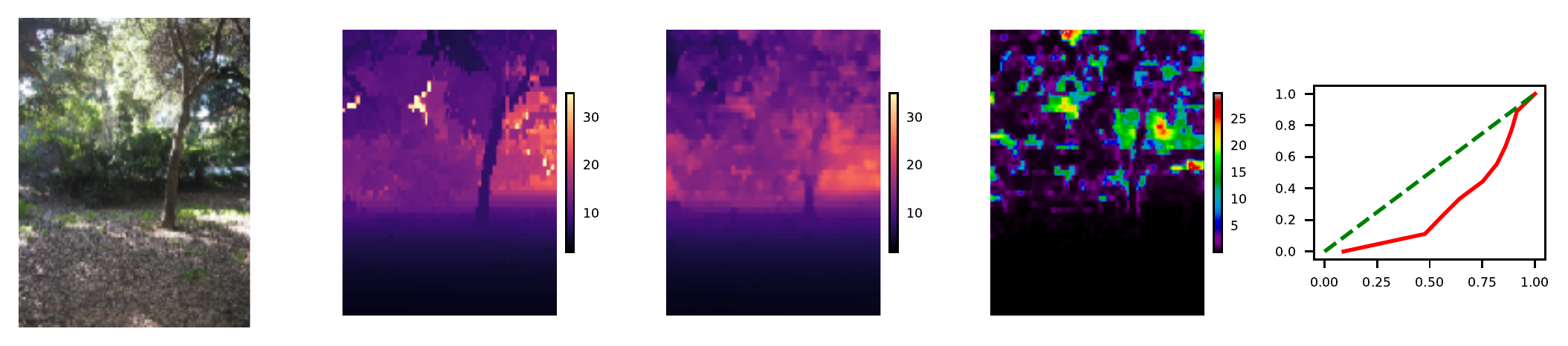} }}%
    \qquad{{\includegraphics[width=15cm]{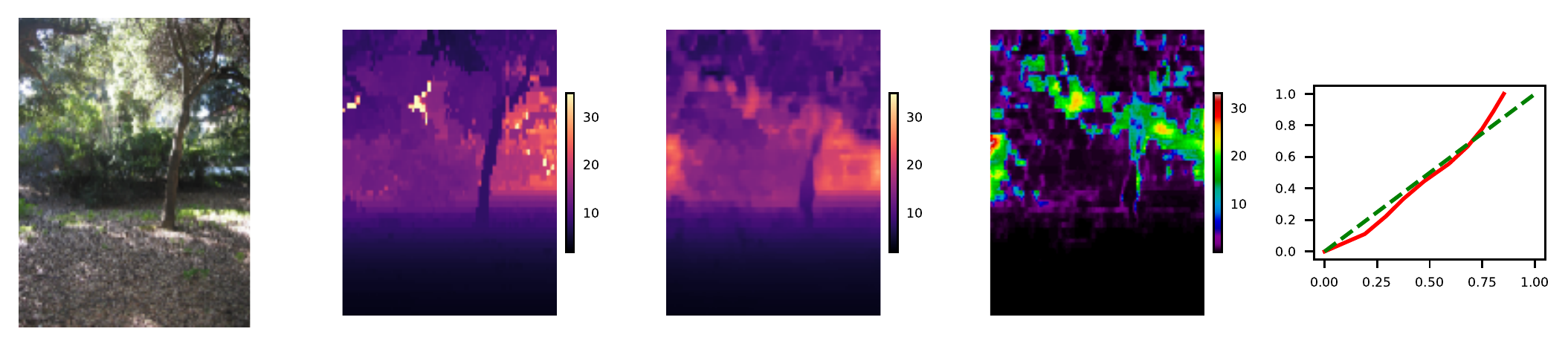} }}%
    \caption{Depth estimation on Make3d. MCDropout-Laplace (first row), Ours-Laplace (second row), Ours-Gaussian (third row), Ours-berHu (fourth row). From left to right: rgb input, ground truth, predictive mean, predictive standard deviation, calibration plot.}
    \vspace{-5mm}
    \label{fig_depth}
\end{figure*}

\begin{figure*}[h!]
    \centering
    \hspace{-7mm}\qquad{{\includegraphics[width=15cm]{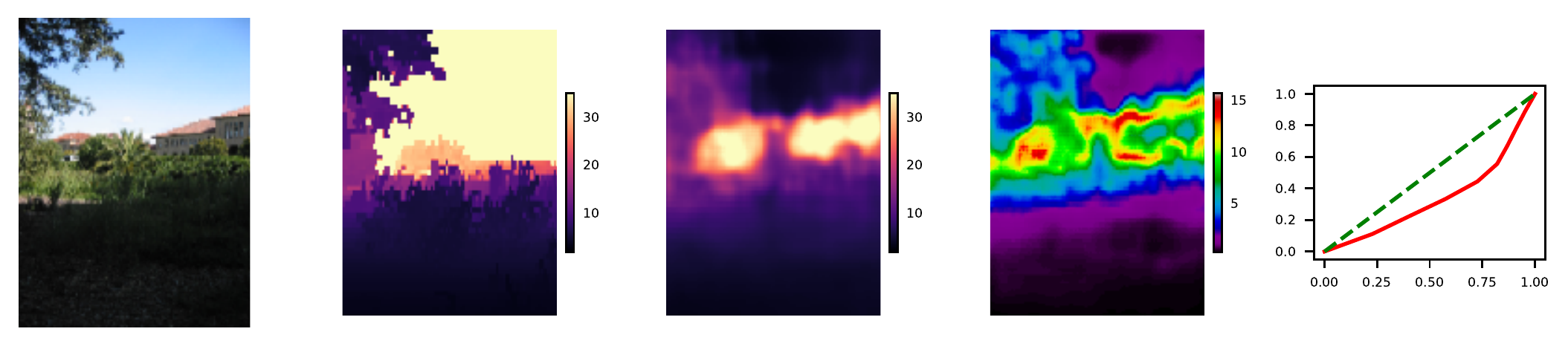} }}%
    \qquad{{\includegraphics[width=15cm]{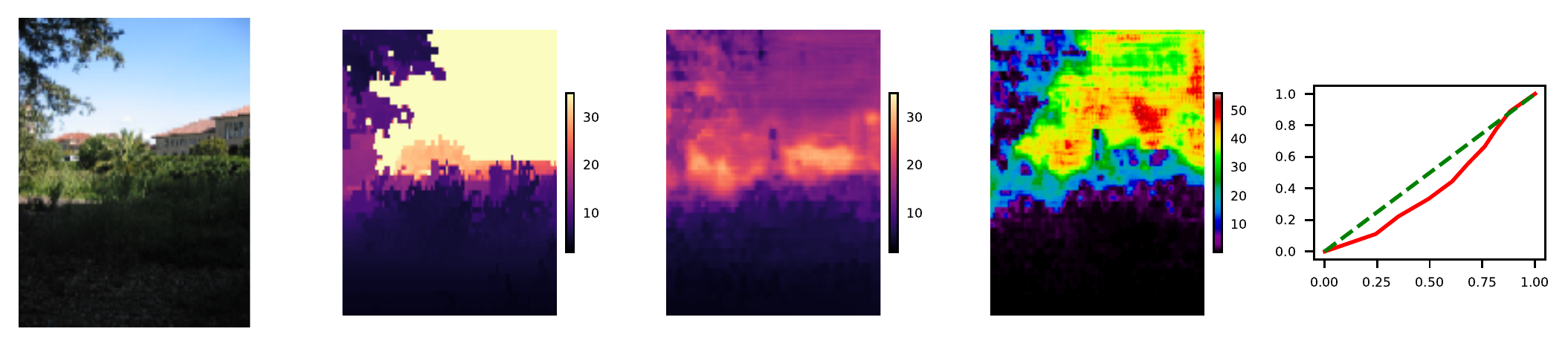} }}%
    \qquad{{\includegraphics[width=15cm]{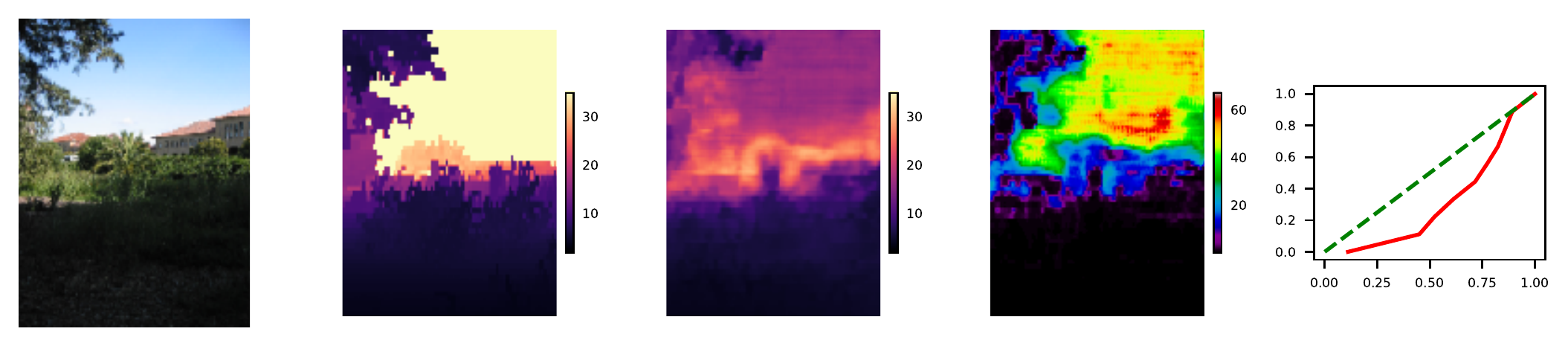} }}%
    \qquad{{\includegraphics[width=15cm]{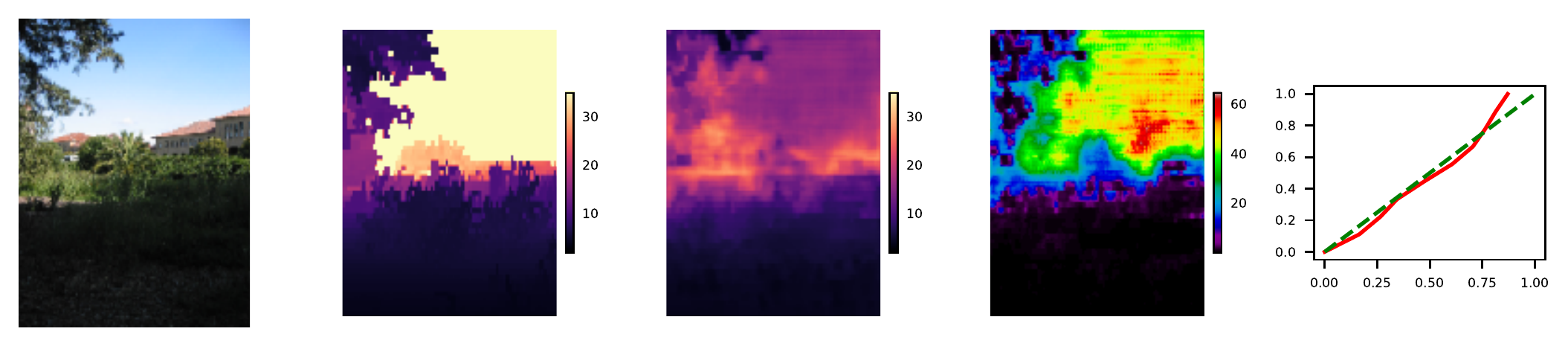} }}%
    \caption{Depth estimation on Make3d. MCDropout-Laplace (first row), Ours-Laplace (second row), Ours-Gaussian (third row), Ours-berHu (fourth row). From left to right: rgb input, ground truth, predictive mean, predictive standard deviation, calibration plot.}
    \vspace{-5mm}
    \label{fig_depth_failure}
\end{figure*}


\end{document}